\documentclass[journal,transmag,screen=true]{IEEEtran}

\ifCLASSINFOpdf
  \usepackage[pdftex]{graphicx}
  \graphicspath{{/figures/}}
\else
  \usepackage[dvips]{graphicx}
  \graphicspath{{/figures/}}
\fi

\usepackage{subfig}
\usepackage{balance}
\usepackage[T1]{fontenc}
\usepackage{booktabs}
\usepackage{dblfloatfix}

\begin{document}

\title{Detection and Analysis of Content Creator Collaborations in YouTube Videos using Face- and Speaker-Recognition}

\author{\IEEEauthorblockN{ Moritz Lode\IEEEauthorrefmark{1}, Michael \"Ortl\IEEEauthorrefmark{1}, Christian Koch\IEEEauthorrefmark{2}, Amr Rizk\IEEEauthorrefmark{2}, Ralf Steinmetz\IEEEauthorrefmark{2}}
\IEEEauthorblockA{\IEEEauthorrefmark{1} Technische Universit\"at Darmstadt, Germany}

\IEEEauthorrefmark{2} Multimedia Communications Lab,
Technische Universit\"at Darmstadt, Germany\\

Email: \{Moritz.Lode | Michael.Oertl\}@stud.tu-darmstadt.de, \{Christian.Koch | Amr.Rizk | Ralf.Steinmetz\}@kom.tu-darmstadt.de

}

\IEEEtitleabstractindextext{%
\begin{abstract}
Abstract---This paper demonstrates the application of speaker recognition for the detection of collaborations in YouTube videos. Here, we extend  CATANA, which is an existing framework for detection and analysis of YouTube collaborations that utilizes face recognition for the detection of collaborators and naturally performs poor on video-content without appearing faces. This work proposes an extension that uses active speaker detection and speaker recognition to improve the detection accuracy in YouTube collaborations.
\end{abstract}
}

\maketitle
\IEEEdisplaynontitleabstractindextext
\IEEEpeerreviewmaketitle

\section{Introduction}

\IEEEPARstart{Y}{ouTube} is one of the most popular video-sharing platforms, being the second largest traffic source among all websites globally~\cite{Alexa2016}. Since Google purchased YouTube in 2007, the YouTube environment changed significantly, with Google implementing the so-called YouTube Partner Program (YPP). As of then, the users were given the possibility to monetize their uploaded videos. YouTube thereby monetizes videos through advertisement and paid subscriptions. Through this, the number of uploaders, or so-called content-creator increased
significantly. There are now more than 10,000 of these content creators making a living solely from online videos~\cite{Youtube2017}. 
Moreover, as of today, YouTube receives video content at a rate of more than 400 hours/min~\cite{Tubularinsights2015}.
Additionally, this brought profit-oriented organizations into YouTube, so-called multi channel networks (MCN) formed. MCNs thereby incorporate up to hundred-thousand channels. Associated channels are supported by the production of video content, for which in return the MCN obtains a share of the channel's revenue.
To increase the popularity of a channel, measured in view and subscriber counts, channels established strategies like collaboration with other channels. A collaboration occurs when two or more channels create a video together and upload this to one or multiple of their channels. This interaction aims to increase the popularity of the channels by potentially combining their audience and is well established and often encouraged by MCNs.
However, no reliable information stating collaborations is available from YouTube. To determine collaborations, it is necessary to detect the interaction in video-content. Through this detection, the effects and properties of the collaborations on YouTube popularity can be analyzed. This knowledge can then be leveraged for improving YouTube strategies, as well as popularity predictions, e.g., for content distribution.

\section{Motivation}
For the detection of collaborations in YouTube, the existing work CATANA~\cite{CATANA} leverages
occurring faces in the video through face recognition to identify the collaborators. 
While CATANA yields good results on video content with a high number of appearing faces, in cases with less or no appearing faces it would not perform well resulting in no detections. One of these cases are videos in the popular category \textit{Gaming}, in which primarily the  game content is visible and often no faces appear. In this prior work, videos of the category Gaming represented a significant part of the used dataset, with $43$\% of all videos.
To improve the overall detection results and cover cases with no visible faces, we propose to leverage the audio content as well. To do this, active speaker detection and speaker recognition are applied to the videos, additionally to the face recognition approach. For this, an extension of the existing framework CATANA is presented in this work.

\section{Related Work}
\label{sec:related_work}
In the following, we introduce existing work on the subject of collaboration detection, specifically the CATANA~\cite{CATANA} framework. Further related work on the applied methods of speaker recognition as well as face tracking and active speaker detection are discussed in this section.

\begin{figure*}[!t]
\centering
\subfloat[CATANA system architecture~\cite{CATANA}.]{\includegraphics[width=\columnwidth]{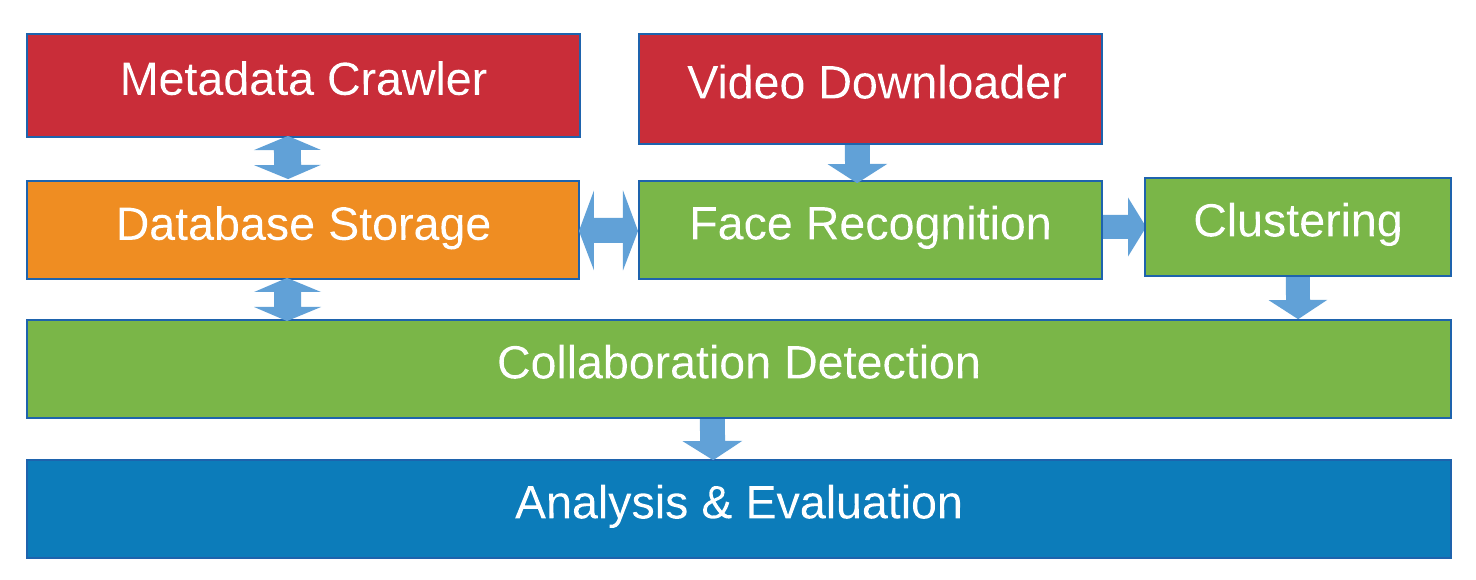}\label{fig:catana_system}}
\hfil
\subfloat[Video views growth factor for collaborations~\cite{CATANA}.]{\includegraphics[width=0.8\columnwidth]{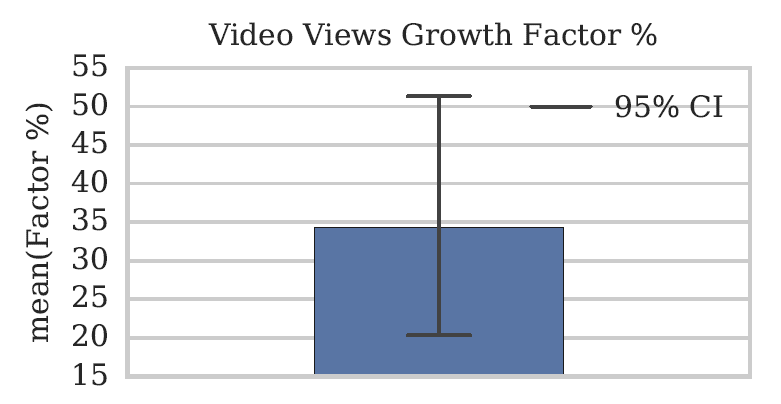}\label{fig:collab_growth_factor}}
\caption{(a) CATANA architecture. (b) Impact of collaborations on the growth of the number of video views. \label{fig:ev_collab_pairwise_nof}}
\end{figure*}

\subsection{Collaboration Detection}
\begin{figure}
\centering
\includegraphics[width=2in]{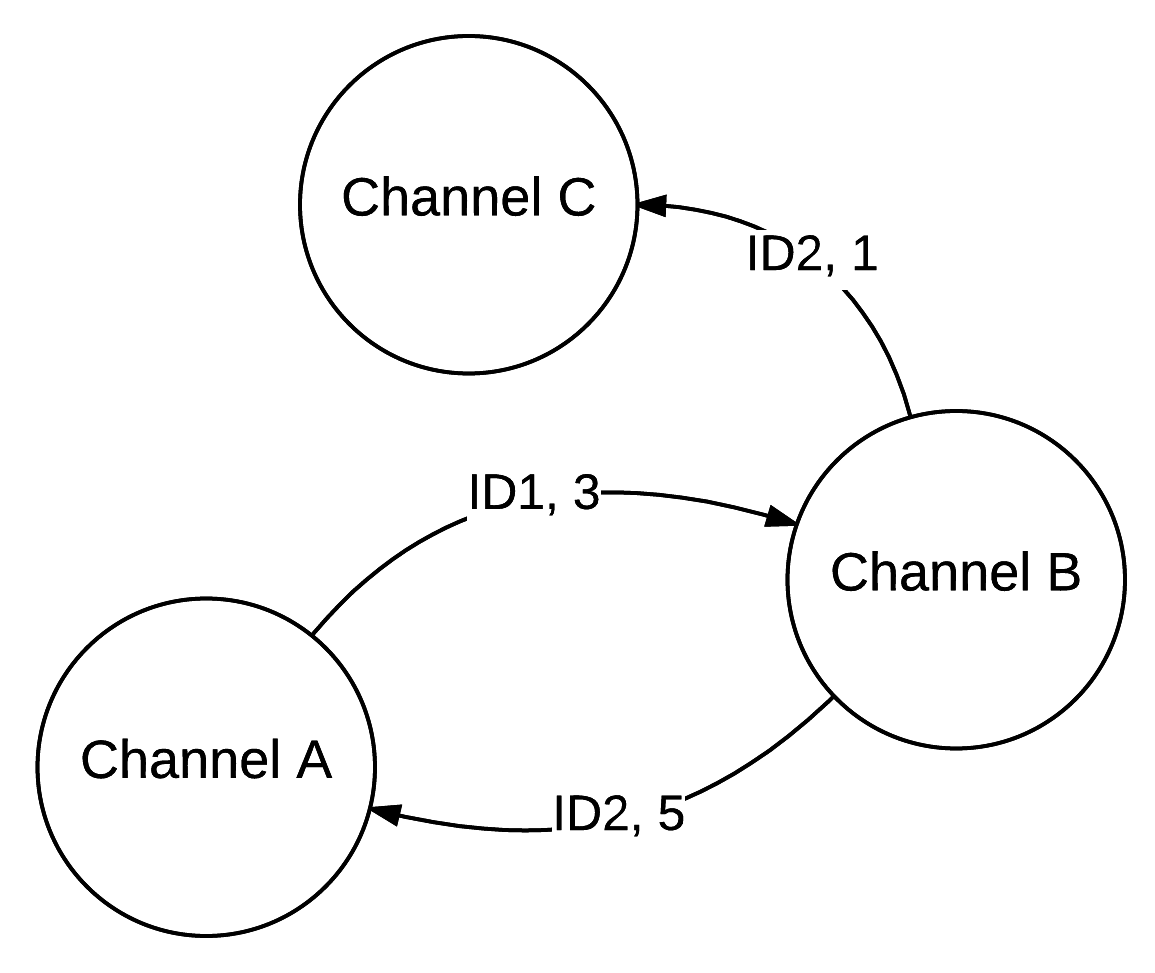}
\caption{Collaboration graph model~\cite{CATANA}.}
\label{fig:fig_collab_model}
\end{figure}

Concerning the analysis and detection of collaborations in YouTube, to the best of our knowledge, not much existing
work was conducted until now.
Koch et al.~\cite{CATANA} proposed an unsupervised framework for collaboration detection in YouTube videos based on face recognition named CATANA. In addition to the detection, the framework is also capable of collecting and analyzing YouTube statistics concerning collaborations and their impact on YouTube popularity. The system architecture is illustrated in Figure~\ref{fig:catana_system}.
For the detection of collaborations in video content, the framework detects the appearing individuals in a set of videos, and creates associations between them. This procedure is similar to a multi-class classification problem, for which no prior knowledge about the number of individuals, nor training data is available.
Individuals are thereby detected frame-wise using an adaptive frame selection method.
Detected faces are then further processed through a CNN-based face recognition method extracting 1,792-dimensional face embeddings which are stored for clustering in a later step. Global clustering on all found face embeddings using the HDBSCAN clustering algorithm~\cite{hdbscan} then associates the appearing individuals across videos.
These associations are leveraged for the assignment of channel content-creator and the collaboration detection based on the number of appearances in a channel.
The found collaborations are thereby modeled as a graph, with channels as nodes and collaborations as directed edges between channels. Figure~\ref{fig:fig_collab_model} shows an example of such a graph for the channels \emph{A}, \emph{B}, \emph{C}, and collaborations concerning the persons \emph{ID1} and \emph{ID2}. A collaboration is thereby modeled as a directed edge. The edge \emph{ID1, 3} for example, describes the appearing of \emph{ID1} from channel \emph{A} in a total of three videos of channel \emph{B}.\\
In the related work \cite{CATANA}, data for a total of 7,492 channels and more than 200k videos were collected in a timespan of 3 months. Due to difficulties in videos without any appearing faces and computation-constraints, only a subset of these videos were analyzed. In total 3,925 collaborations for 1,599 channels were found in this time, on average 2.8 collaboration per channel. Further different aspects of the found collaborations concerning categories, popularity and MCN-memberships were analyzed. Results show that collaborations occurred mostly in the category Entertainment. Concerning MCNs membership, collaborations within their own network, or with non-associated channels were preeminent. Concerning the popularity impact of collaborations, a positive effect in both channel and video statistics was found. One of them is the average growth of collaboration video's view-counts, which was found in average 34\% higher than these of non-collaboration videos of the same channel. Figure~\ref{fig:collab_growth_factor} shows the mean video view-count growth factor using a confidence interval of 95\%.\\
Difficulties in the detection of collaborations in video content without appearing faces were found, especially in the prominent video category Gaming, which constituted 43\% of the total number of videos in their dataset.\\
Due to these difficulties with nearly half of their video data resulting in incorrect or no detections, a more sophisticated approach improving the detection rate is necessary. For this reason, we will build on the existing framework and extend the detection by using active speaker detection and speaker recognition.

\begin{figure*}[!t]
\centering
\includegraphics[width=0.8\textwidth]{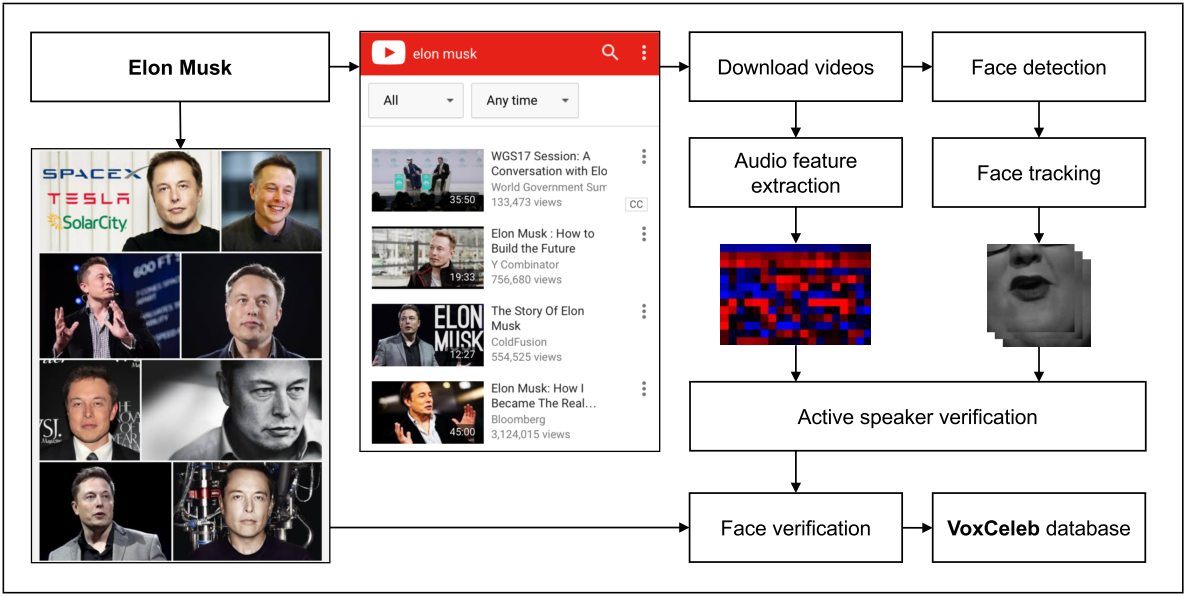}
\caption{VoxCeleb active speaker verification pipeline as given in \cite{Nagrani2017}.}
\hfill
\label{fig:vggvox_system}
\end{figure*}

\subsection{Speaker Recognition}
In this section, relevant existing approaches on speaker recognition are presented and the possible application in this work discussed.\\
Nagrani et al.~\cite{Nagrani2017} present VoxCeleb, in which they made multiple contributions concerning speaker recognition. First, a fully automated pipeline is proposed for the creation of a speaker recognition dataset from online videos.
The pipeline thereby combines active speaker detection and face recognition to identify a speaker in the video content and is illustrated in Figure~\ref{fig:vggvox_system}. For a predefined list of $2,622$ people, videos from YouTube are extracted and analyzed through the pipeline. Finally resulting in utterances of $1,251$ people. Reason for the difference in the number of individuals is that during the pipeline process videos of individuals were discarded if no active speaker could be detected or the speaker not identified. 
Using existing work of Chung et al.~\cite{chung16} for active speaker detection and Simonyan et al.~\cite{Simonyan15} for face recognition, hundreds of thousands of video-audio segments were automatically extracted with a high precision concerning speaker identity.
This collected speaker dataset called VoxCeleb~\cite{Voxcelebdataset} is made publicly available and consists of over $140$k utterances of $1,251$ individuals. The dataset was then further leveraged in their second contribution, a CNN-based architecture for speaker identification and verification.
This architecture is based on the work of VGG-M~\cite{Chatfield14}, which is a network designed for image classification achieving state of the art performance in this task. The image classification architecture is thereby leveraged by generating spectrograms in a sliding window fashion, each representing $3$ seconds of speech, and uses these as input.
The proposed network was then trained using the described VoxCeleb speaker dataset.
For the speaker identification task a $1,251$-dim. softmax layer is used as the output to produce a classification over the 1,251 individuals in the dataset. For identification, an accuracy of $92.1$\% was achieved. Additionally, they compared their approach with the de facto state of the art in the recent years, an i-vector based~\cite{Dehak11} approach. This approach thereby achieved only an accuracy of $75.6$\% on the dataset.

For the verification task, feature vectors are extracted one layer before the softmax output, and leveraged as a speaker embedding. This $1,024$-dim. embedding can be directly used for speaker comparison using a distance metric like the cosine distance. Verification evaluation on the dataset using Equal Error Rate (EER) resulted in $7.8$\% for the CNN-based approach, while only $8.8$\% for i-vectors (the less, the better).\\
The speaker recognition network is made available in the form of pre-trained models and thus can be directly used in this work. The proposed embeddings for speaker verification can thereby be leveraged in classification, as well as clustering. Especially clustering is of importance in this work, as we do not have prior training data for classification.

\begin{figure*}[!t]
\centering
\includegraphics[width=0.5\textwidth]{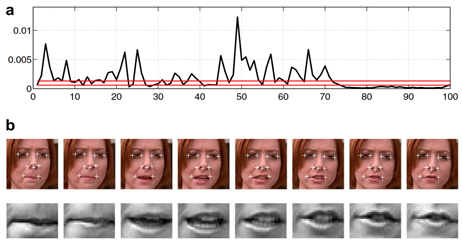}
\caption{Active speaker detection sample using the inter-frame difference of the mouth region. The inter-frame difference (a) is bigger than the threshold while the person is talking. At frame 73 the person stops talking and the inter-frame difference is below the threshold~\cite{everingham09}.}
\hfill
\label{fig_collab_model}
\label{fig:interFrameDifference}
\end{figure*}

\subsection{Speaker Diarization}
Speaker diarization is the task of dividing an audio sample into segments of different speakers.
The task thereby consists of multiple steps, voice activity detection (VAD), speaker model generation, and clustering. Different existing approaches are available and discussed in this chapter.\\
Rouvier et al.~\cite{Rouvier2013} present the LIUM (Laboratoire d'Informatique de l'Universit\'{e} du Maine) Speaker Diarization, an open-source toolbox for speaker diarization in news broadcasts.
The toolbox is developed to provide a tool for the development of new diarization systems, or the direct application as an easy to use speaker diarization tool.
The tool thereby computes the Mel Frequency Cepstral Coefficients (MFCC) parameter from the audio and provides pre-trained universal background models (UBM) for the classification of speech, non-speech segments.
In the following either hierarchical agglomerative clustering using cross-likelihood ratio (CLR) or integer linear programming using i-vector~\cite{Dehak11} clustering is available for separating the speaker.
The proposed system is evaluated on three datasets, ESTER\footnote{http://catalog.elra.info/product\_info.php?products\_id=999 [\today]}, ETAPE\footnote{http://catalog.elra.info/product\_info.php?products\_id=1299 [\today]}, and REPERE\footnote{http://catalog.elra.info/product\_info.php?products\_id=1241 [\today]}.
Evaluation metric is the diarization error rate (DER) to measure the performance. DER thereby
measures the speaking time attributed to the correct speaker and was introduced by NIST~\cite{NISTDER}.
Results show a DER of $8.35$\% - $24.49$\% for the three evaluated datasets.
A direct comparison with other existing diarization methods was however not conducted.
With this insights, the decision for applying this toolbox in our work should be based on further tests
on YouTube video content, as this toolbox is mainly dedicated for news broadcasts.\\
An additional approach concerning speaker diarization in this work can be achieved by combining different existing components to a diarization pipeline consisting out of the parts of voice activity detection, speaker modeling, and clustering.
For speaker modeling, the already discussed VoxCeleb speaker embeddings model~\cite{Nagrani2017} can be used to model a speaker to a feature embedding, which then can be directly applied in clustering.
For the voice activity detection task multiple existing tools are available. 
One of the current state of the art VAD systems is developed by Google for the WebRTC~\cite{WEBRTC} project, which is freely available and easy to use. WebRTC is an open-source framework for Real-Time Communications (RTC) on the web, consisting of high-quality components for online communication such as voice and video chat~\cite{WEBRTC}. The VAD system supports segments of $10$, $20$, or $30$ms and uses multiple frequency band features with a pre-trained Gaussian mixture model (GMM) for classification~\cite{Salishev2016}.
By applying a VAD system, an audio sample is segmented into speech, non-speech segments. 
To construct a speaker diarization system from these components, an audio sample is first divided by the WebRTC VAD and in the following the segments applied to the VoxCeleb network extracting speaker embeddings. For these extracted speaker embeddings a clustering algorithm using a metric, for example, the cosine distance, is then applied to differentiate the occurring speakers.

\subsection{Face tracking}
In order to incorporate the speaker recognition into CATANA, it is necessary to switch from a frame-based face detection to a face tracking approach. The resulting face tracks are a representation of a person occurring in the video in several sequential frames. This representation further makes it possible to draw inferences about the relation to the audio track, which is analyzed by the speaker recognition. Everingham et al.~\cite{everingham09} discussed the differences between a frontal and a multi-view face detection. Although the frontal face detector \cite{mikolajczyk} allows only to detect frontal faces, it also is more reliable than multi-view face detectors such as \cite{li04}.

\begin{figure*}[!t]
\centering
\subfloat[Visualization of the interesting face points and how they move in 60 frames. Frame $0$ and $60$ are represented as the original frame.]{\includegraphics[width=0.8\columnwidth]{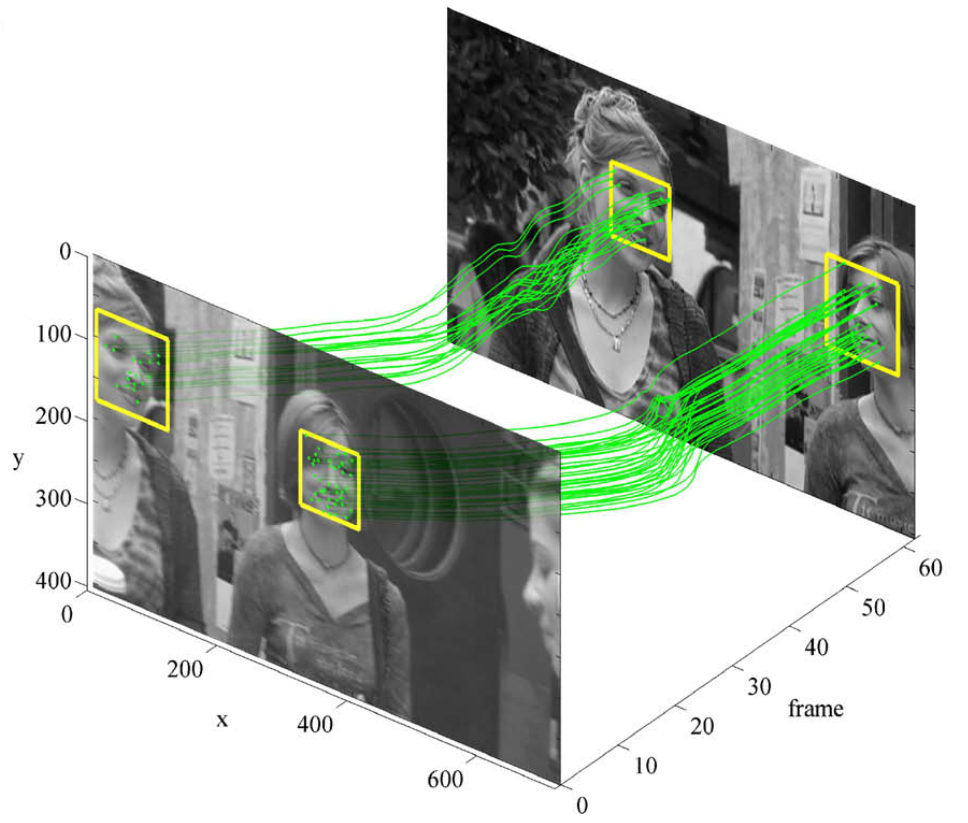}\label{fig:facetrack}}
\hfil
\subfloat[As a preprocessing step, the audio layer has to be converted to a heatmap.]{\includegraphics[width=\columnwidth]{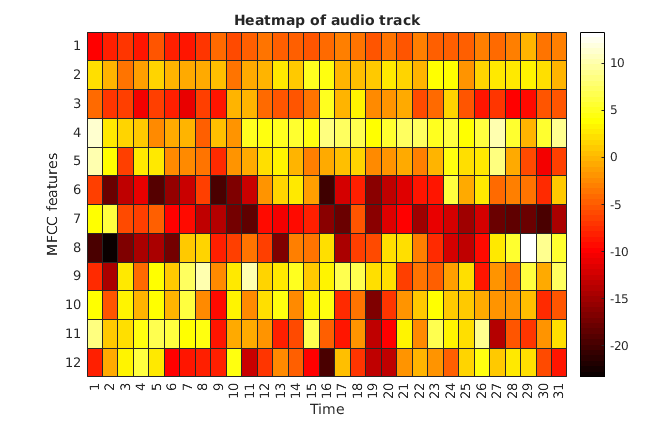}\label{fig:heatmap}}
\caption{Face tracking in frames and audio input converted for active speaker detection.\label{fig:face_tracking}}
\end{figure*}

A popular algorithm to track faces is the Kanade-Lucas-Tomasi (KLT) tracker. The tracking is implemented by matching interesting points of the detected face to the local area of the following frames. The output is a stream set of interesting points for each frame. This method is visualized in figure \ref{fig:facetrack}. It is robust against camera movement and is also capable to track moving persons. Especially compared directly to frame-based face detection, this method can detect face tracks although the face is not visible continuously. Also, the computational effort of the KLT is less than applying face detection on every frame.

The VGG (Visual Geometry Group) Face Tracker \cite{VGGFaceTracker} is a MATLAB toolbox for face tracking provided by the University of Oxford. It comes with a pre-trained cascade face detection model, which makes it possible to start right away without the need to train a model yourself. The VGG Face Tracker works in four basic steps. First, the frames are extracted from the video using the FFmpeg software~\cite{ffmpeg}. The second step is to detect the shot boundaries by color thresholding. For the third part, the face detection, Face Tracker uses the pre-trained model. 

The last segment of the toolbox is the tracking functionality, which uses the KLT algorithm. The big advantage of this toolkit is that it can be configured and adapted for every use case. Either fast tracking with less accuracy or highly reliable tracking with a high amount of computational effort, both is possible. We try to make use of both worlds.

\subsection{Active Speaker Detection}
There are several methods for implementing active speaker detection. Prerequisite is a preprocessed video labeled with face tracks. An intuitive method for speaker identification is by detecting lip movement directly in the face-images of the track. Everingham et al.~\cite{everingham09} proposed to calculate the inter-frame difference of the corresponding mouth regions of the detected faces. This measurement is calculated by taking the sum of the squared difference of the pixel values. As shown in Figure~\ref{fig:interFrameDifference}, the inter-frame difference of the mouth region is higher if the person is speaking. However, this method may not be correct in all cases, for example, if the person moves the mouth without speaking, e.g., while eating or laughing.

Chung et al.~\cite{chung16} proposed the audio-to-video synchronization network (SyncNet). It is a two-stream convolutional neural network (CNN), which consists of two CNNs, one for the audio and one for the video track. Initially, the authors created the SyncNet for audio-visual synchronization tasks, which are very important for the film industry, for example, when the audio part is streamed separately from the video part. But the SyncNet can also be used for active speaker detection because it calculates how well the audio fits to the mouth region of a person. The network needs two inputs. First, it needs $0.2$ seconds of the audio stream, which needs to be converted to a heatmap. The heatmap is illustrated in Figure \ref{fig:heatmap}. The $x$-axis thereby represents the MFCC values for each time step and on the $y$-axis the color indicates the power if the frequency bins. The video needs to be converted to a $25$Hz frame rate and cropped to the face region with $160$x$160$ pixels.  SyncNet can be used with the MATLAB toolbox MatConvNet \cite{MatConvNet}. It also comes with a pre-trained model, which was trained on BBC videos with several hundreds of speakers.
The network calculates a confidence of how likely it is for the mouth region to be part of the speaking person. So the speaker will most likely be the one with the highest confidence value. On the Columbia dataset \cite{chakravarty16}, the work achieved an accuracy between $99.8$\% and $100$\%.

\section{Design}
We propose an extension of the existing CATANA framework through four additional methods: face tracking, active speaker detection, speaker diarization, and speaker recognition.

For face detection, a frame-wise approach is used in CATANA. To get more information concerning the temporal
course of the appearances we choose to apply face tracking instead. Face tracking leveraged shot boundary detection to extract a sequence of subsequent frames of appearing faces. Through the temporal dimension of the sequence, a direct correlation between the frame-sequence and its audio can be made. 

In the following, active speaker detection is applied to decide wherever the detected face in the sequence is the speaker of the
associated audio. If the active speaker is visible, the previous steps provide both face and speaker information, which can then be later used combined or separately to identify the person.
If none of the visible faces is associated as the speaker, face recognition is applied on the detected faces like in the existing CATANA approach nonetheless.
If no visible speaker is detected, speaker diarization is applied to potential detect and separate the
non-visible speaker in the audio. These separated speech segments are applied through speaker recognition to identify
a speaker without face information.
\begin{figure*}[t]
\centering
\includegraphics[width=0.6\textwidth]{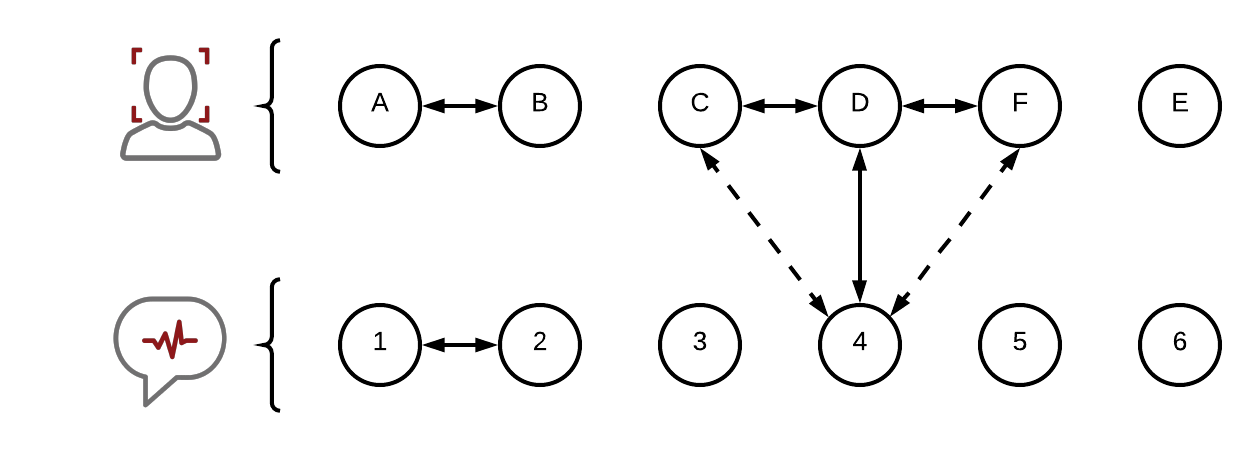}
\caption{Face and speaker embeddings bridge.}
\hfill
\label{fig:fig_face_speaker_bridger}
\end{figure*}

The result of these steps is either a combination of face and speaker embeddings, describing the visible active speaker, only face embeddings of appearing faces, or only speaker embeddings describing a non-visible speaker, for example, a commentator.
Especially the first case, for which both types of embeddings are available is most helpful in later identifying the persons appearing. The availability of both embeddings for a segment of speech makes it possible to bridge
face and speaker embeddings and thus associate a face with a voice. 
The goal is thereby to gain information by associating stand-alone face or speaker embeddings with the found pairs of embeddings and thus be able to identify, for example, a non-visible speaker in videos.
We achieve this by first clustering the found face and speaker embeddings separately, which results in associations between face-to-face and speaker-to-speaker embeddings. 
For pairs of face and speaker embeddings found for an active speaker, an additional association between the different types of embeddings exists, which is further extended to embeddings clustered within their type.

Figure~\ref{fig:fig_face_speaker_bridger} shows a schematic of this process, modeling embeddings as circles and edges as an association between them. Displayed at the top are face embeddings and at the bottom the speaker embeddings.
Face embedding \emph{D} thereby has an association with speaker embedding $4$, describing an active speaker detection. Through the separate clustering of the face and speaker embeddings, associations within their respective types were created. An example are the embeddings \emph{C}, \emph{D}, and \emph{F}. Through this, implicit associations can further be made between the embeddings \emph{C}, \emph{F} and $4$, as \emph{C}, \emph{D}, \emph{F} model the same face and thus same person.

\section{Implementation}
\begin{figure*}
\centering
\includegraphics[width=0.6\textwidth]{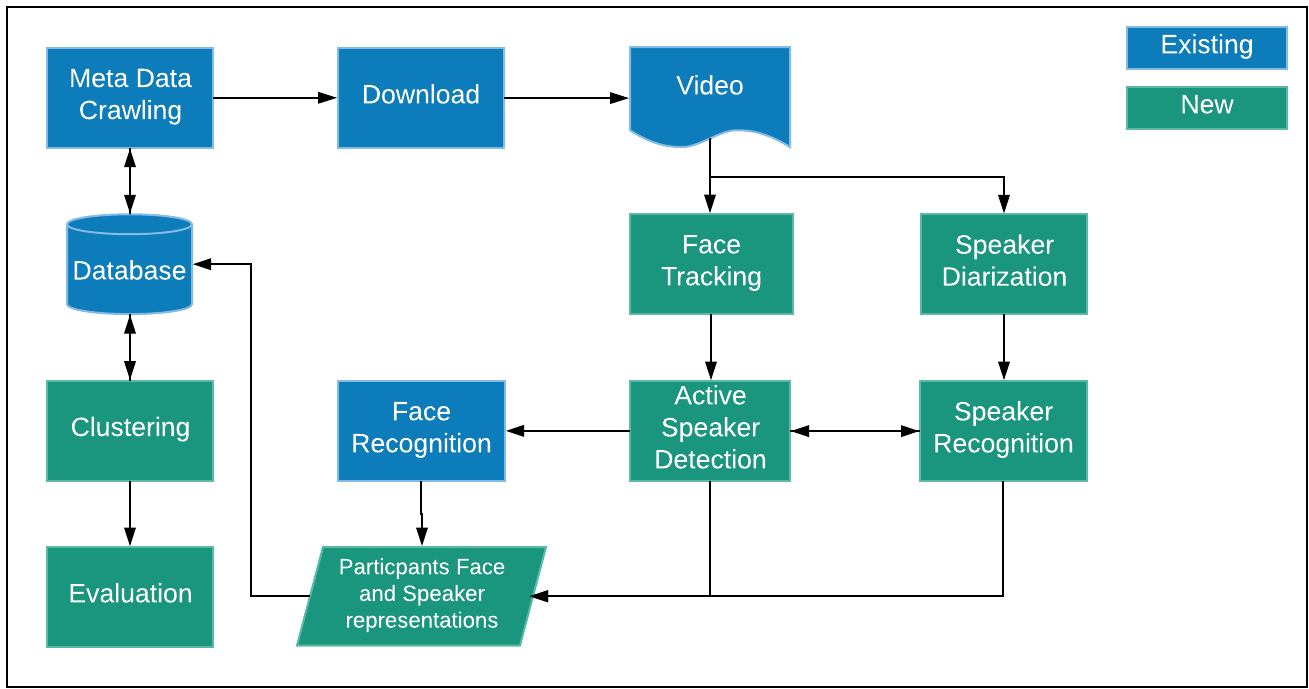}
\caption{The proposed extension of the CATANA architecture. Blue segments are reused modules from CATANA, green segments show added or updated parts.}
\hfill
\label{fig:proposed_system}
\end{figure*}

Our implementation is based on the CATANA framework, which is shown in Figure~\ref{fig:catana_system}. Our proposed extension is illustrated in Figure~\ref{fig:proposed_system}. The changed parts of the system are highlighted and labeled, wherever an existing CATANA part is used unchanged, or replaced with a new implementation. Color blue describes parts of the CATANA framework which were left unchanged like data storage, metadata crawler, and video download. Green describes updated or replaced parts. Prominently the former face recognition and clustering part, in which the video is processed, were replaced for the most part. Instead of only applying face recognition, both active speaker detection and speaker diarization are executed before face recognition, returning potential speaker-face associations.
The extension is divided into three major parts: face tracking, speaker recognition, and clustering.\\
The whole CATANA framework is implemented using Python, which we will continue using to extend the framework. However, some parts of the proposed extension are using a different runtime environment and therefore require an interface to Python. VGGVox, VGG Face Tracker, as well as SyncNet are implemented using MATLAB. Due to incompatibility, it is not possible to use the pre-trained models directly in Python, making the interaction between Python and MATLAB necessary. To accomplish this the MATLAB Engine API for Python~\cite{matlabAPI} is used. This library provides an official interface between Python and MATLAB. It is possible to call MATLAB functions, scripts, provide parameters, and receive return values. In practice, a MATLAB instance is started as a background process and executes all calls from Python. Data interaction between Python and MATLAB thereby only supports a set of compatible types which are converted between equivalent types of both technologies. For this reason, multiple of the MATLAB function's return values needed an additional adjustment in terms of type and structure, i.e., only scalar structures with size 1 can be returned.

\subsection{Face Tracking and Active Speaker Detection}
\label{face_tracking_and_active_detection}
These two steps are implemented with MATLAB. The VGG (Visual Geometry Group) Face Tracker \cite{VGGFaceTracker} is used for the face tracking. It is required to install FFmpeg \cite{ffmpeg} and the Matconvnet \cite{MatConvNet} toolbox to run the algorithm. We have adapted the algorithm to our needs to provide a faster detection, but it remains still very time consuming, especially if the video is long. The first two steps are frame extraction and shot detection, which work quite fast, compared to the third and fourth step, face detection and face tracking, which tend to take a while, because a large number of frames must be processed. We decided to apply the face detection only on every $25$th frame, which is equivalent with one detection per second. People who occur only inside this time frame will not be considered. We assumed that if a person appears less than a second, it does not contribute enough to the video to be considered. This speeds the process up and in advance filters the faces which are not of interest. We could have raised the frame skip, but then the probability of missing a face is higher and the benefit of that does not have such a big impact. The last step could not be made faster, because the face tracking of the face has to be computed on every frame. This is especially necessary if the observed person is moving. The output of the face tracking is the location information for every found face in every frame.
To make sure the face tracks will not get too long, they are cut into pieces with the length of $50$ frames. 
This is done to ensure, that tracks where a person starts and stops speaking are separated, so that the evaluation of the person being the active speaker can be more accurate. We made the decision to cut the tracks down to $50$ frames, because a two-second audio stream is sufficient to extract speaker embeddings. If the confidence of SyncNet is high enough we calculate the embeddings. We also defined a minimal length of the face tracks. This is important, because the input for the speaker embedding needs to be long enough to provide good and reliable results. Afterwards, every face track is stored to the hard drive as an \emph{.avi} video file with the corresponding audio track. 
We decided to use the SyncNet provided by Chung et al.~\cite{chung16} for the active speaker detection, because of their good results and high accuracy. On the Columbia dataset~\cite{chakravarty16} they reported F1-scores between $83.4$\% and $97.7$\% for a frame window of $10$. By raising the window to $100$ they reported F1-scores between $99.8$\% and $100$\%. They provide a neural network to calculate the confidence if a person is speaking. The network thereby needs a video file of the cropped face with the audio track as input. Before the video can be further processed it needs to be converted into the right format. The algorithm can only work with a video at 25 frames per second (fps) and an audio sampling rate at 16 kHz. 
We made sure that the cropped images are as similar as possible to the provided examples, as the neural network performance is depended on the similarity of the input and its training data. In order to achieve this similarity, the mouth of the person is placed in the center of the picture and it is resized to a size of $160$x$160$ pixels. After preparing all necessary steps and preprocessing the data we determined that the output of the SyncNet network is not as desired and seemed to be highly random. Eventually after debugging and different experiments with video and audio codecs we found that it is not enough to convert the video to an $.avi$ file with $25$ fps and a sampling rate of $16$ kHz, the audio codec needs to be PCM S16 LE (araw) and the video codec needs to be in $24$ bits RGB (RV$24$). After this insight, we could process with the further steps.\\
For tracks with a high confidence that the person is speaking we directly calculate a speaker embedding. As we now obtained both speaker embedding and face images, which are applied to face recognition, we can now identify a person by its voice and its face separately, if the person is visible and speaks at the same time at least once, and if the active speaker recognition calculates a high confidence.

\begin{figure*}
\centering
\includegraphics[width=0.95\textwidth]{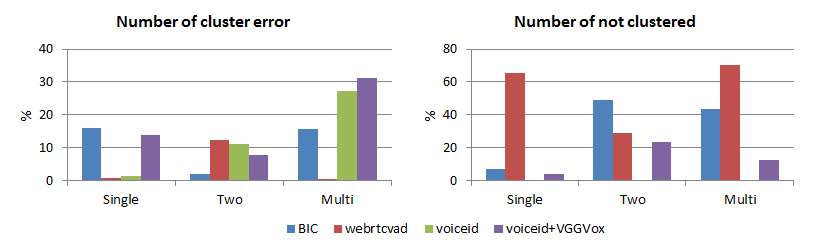}
\caption{Diarization evaluation for number of incorrect clustered segments and not clustered segments.}
\label{fig:diarization_eval1}
\end{figure*}
\begin{figure*}
\centering
\includegraphics[width=0.95\textwidth]{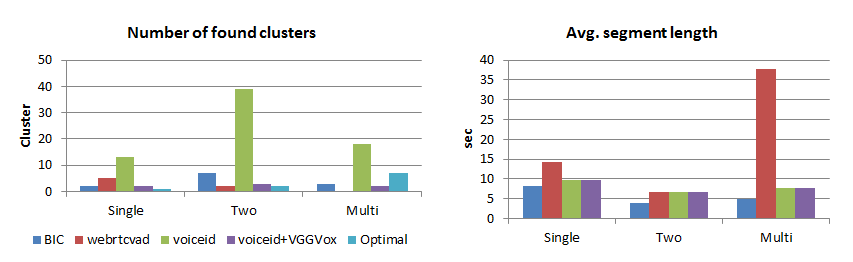}
\caption{Diarization evaluation concerning number of clusters, also displaying the optimal number for the respective class.}
\label{fig:diarization_eval2}
\end{figure*}

\subsection{Speaker recognition}
The speaker recognition parts of our pipeline extension consist of speaker recognition for identifying different speakers and speaker diarization, which segments audio into speech and non-speech segments.\\
For speaker recognition the proposed speaker embedding system VGGVox~\cite{Nagrani2017} is used. The system is developed in MATLAB and also requires the additional toolbox Matconvnet~\cite{MatConvNet}. The system expects an audio sample as input and returns a $1,024$-dim. float vector describing the speaker present. The audio sample thereby requires to have a sampling rate of $16$ kHz, $16$ bit depth and to be single channel (mono).
While testing, we also discovered that a too short audio sample could not be processed through the model, a minimum length of $1$ sec or $16,000$ samples were required to extract an embedding without error.
Speaker diarization is implemented as described in section \ref{sec:related_work} by combining multiple available systems to a diarization pipeline. Before implementing this pipeline we further conducted an evaluation regarding the used parts for diarization and their performance on a set of YouTube videos. For this we defined three classes of YouTube videos based on the number of appearing speakers in the audio. Classes are \emph{Single}, if they consist only of one speaker, \emph{Two} with two speakers, or \emph{Multi} consisting of 4 to 8 speakers. We evaluated on the systems BIC~\cite{dia_BIC}, webrtcvad~\cite{dia_webrtc}, voiceid~\cite{dia_voiceid}, and voiceid+VGGVox.
BIC is a speaker diarization library based on Python performing VAD, audio segmentation and hierarchical clustering.
Webrtcvad is the already described VAD system developed by Google for the WebRTC standard. As webrtcvad is only a VAD system, we combined it with the VGGVox~\cite{Nagrani2017} speaker recognition and HDBSCAN~\cite{hdbscan} clustering to a complete diarization pipeline.
Voiceid is a Python based speaker identification system based on the previously described LIUM Speaker Diarization framework~\cite{dia_voiceid}.
Additionally to the full speaker diarization system of voiceid (LIUM), we also combined voiceids's speech segmentation with state of the art speaker recognition and clustering approaches using VGGVox and HDBSCAN~\cite{hdbscan} forming a new integrated system.\\
Results in Figures~\ref{fig:diarization_eval1} and \ref{fig:diarization_eval2} show different metrics concerning the clustered audio segments: \emph{Number of cluster error} is describing the number of speech segments incorrectly clustered and \emph{number of not clustered} describes the segments not clustered at all. Further the \emph{number of found cluster} displays the number of found speaker cluster, additionally showing the class' respective optimal number. The \emph{average segment length} describes the average length in seconds of the extracted segments for the systems.
We see that for all classes voiceid extracts a large number of clusters compared to the optimal number.
This indicates that their applied clustering algorithm is conservative in merging found speaker and thus yields a lot of small clusters. In general, voiceid+VGGVOX yields the best results for number of clusters. However, a special case is the \emph{Multi} class, for which we discovered that each of the tested systems seems to struggle in distinguishing between a increased number of people. Particular in a discussion, where multiple speakers are talking into one another. This is due to the fact that the VAD system uses the silence between speeches to determine segment boundaries. In a discussion, often no pause is present, a segment can therefore contain multiple speakers which makes it hard to correctly extract an embedding and cluster. This case is also reflected in the number of cluster errors for the \emph{Multi} class.
For the average segment length in seconds, all systems yield similar results, around $6$ seconds in average for classes with more than one speaker, and nearly $10$ seconds for the \emph{Single} class. This shows that in a setting of only a single speaker, speeches seem to have a longer duration between pauses than in a dialogue between multiple speakers.
Noticeable is the spike for webrtcvad in the \emph{Multi} class illustrating the described behavior of detecting no pause, which will likely contain multiple speakers per segment.
We can also see that the webrtcvad system seems to have a low number of cluster error for all classes, even \emph{Multi}.
This is however a false impression, as studying the number of not clustered, we can see that webrtcvad in fact has not clustered the majority of the segments at all.
Concerning these results we decided to further leverage the voiceid+VGGVox approach as our speaker diarization system, as it yielded comparable performance as voiceid but also clustered the segments to the near optimal number of clusters.

\begin{table*}[!t]
\centering
  \begin{tabular}{ l | l  l  l  l l} 
  \toprule
     VoxCeleb 
     Test Set& 	Correct & Incorrect & Homogeneity & Completeness & V-Measure \\ 
     \midrule
    CATANA 	& 		$86\% $ 	&  $14\% $ &  $0.87$	& 	 $0.89$		&	$0.88$	\\ 
    
    Speaker Recognition &  $37\%$ &  	$63\%$	& 	$0.43$	& 	$0.65$	& $0.51$\\
    
    Proposed method & $60\%$ &  	$40\%$	& 		$0.54$	& 		$0.55$		& $0.55$ \\
    \bottomrule
  \end{tabular}
	\caption{The results on the VoxCeleb set for CATANA, speaker recognition and our approach.} 
    \label{table:voxCelebComparison}
\end{table*}

\subsection{Clustering}
Clustering is applied in several parts of the pipeline. In general for all applications of clustering the
HDBSCAN algorithm is used. 
Through HDBSCAN's density based approach, no parameter for the number of clusters or distance i.e. epsilon parameter is necessary, while still yielding comparable, or better results than other approaches~\cite{CATANA}.
As described in the prior work of CATANA, HDBSCAN has problems when only a single class is present in the data, yielding only the noise label. For this reason, we establish a fallback method, like proposed in CATANA, using DBSCAN for clustering, in case only noise was found.
As described in the previous sections, clustering is applied in multiple parts, for example, speaker diarization, speaker recognition, and face recognition which all use clustering at some point to group the extracted data. Additionally, at the end of the pipeline, for the actual collaboration detection, all found embeddings are clustered again globally to associate face or speaker across videos and channels.
For global clustering a Cython\footnote{http://cython.org [\today]} based implementation of the distance matrix computation is adopted from CATANA.
The embeddings are thereby stored using the existing data storage backend of CATANA, consisting of a MySQL database.
To accompany the extension, the original database scheme was adapted to additionally store both face and speaker embeddings.\\
A special case for the application of clustering is the merging of face tracks. This is necessary as too long face tracks are split up as described in section \ref{face_tracking_and_active_detection}, resulting in a large number of tracks which may only contain one individual. To simplify the later storage, tracks of the same individual should be merged back together. This is accomplished by clustering all split-up face tracks based on their face embeddings and merging them according to the found cluster label. For tracks also containing speaker embeddings, this pair of embeddings is treated as one entity and not separated by merging.

\section{Evaluation}
Evaluation on the analyzed YouTube dataset of CATANA is desirable. It would be possible to compare the created collaboration graphs and their properties like number of edges, nodes, and clustering index. However, no labeled data is available to evaluate the actual accuracy of both methods, thus we could not determine if the new proposed method is performing better or not.
For this reason, we have chosen to evaluate on two different datasets. First, we evaluate the previously presented VoxCeleb speaker dataset.
In this dataset the speaker is always visible and labeled data is available.
This allows us to apply both, CATANA and our proposed extension to the data. CATANA thereby only uses appearing faces for classification, while our approach can also leverage speech data. Through this we can determine if speaker recognition can improve the classification accuracy.
The VoxCeleb dataset consists of over 20,000 videos of 1,251 individuals. Due to time constraint of this work we will not analyze the complete dataset but only a small test set of 40 speakers and 700 videos (selected randomly).
Evaluated will be which person is occurring in which video, without any prior training, or knowledge of included individuals (open-set classification scenario).\\
Evaluation metrics used for this first dataset are the percentage of correct and incorrect assigned videos, as well as the clustering metrics homogeneity, completeness and v-measure, as its ultimately a clustering task.
The homogeneity score is a value between $0.0$ and $1.0$, while $1.0$ stands for a perfect homogeneous result. Homogeneity is satisfied if all clusters contain only data points of a single class~\cite{scikit_clustering}.
Completeness, also a score between $0.0$ and $1.0$, describes if all members of a given class are assigned to the same cluster~\cite{scikit_clustering}.
V-Measure is a combined metric of these two and formed by the harmonic mean of the homogeneity and completeness score.\\
The second dataset is a set of YouTube videos collected in this work. Videos were thereby selected based on properties prior difficult for the CATANA framework, like gaming videos with no visible faces. A set of 72 videos of 9 channels was collected.
Both videos with and without collaborations are present. There is a total of 34 collaborations between these channels.
Additionally, only 25 of these 72 videos have appearing faces, while the rest only consists of non-visible speakers.
Goal for this dataset is the evaluation of collaboration detection results for both systems. With particularly attention on the difficult cases of CATANA, we evaluate if our proposed extensions improve the detection results.
Evaluated will be the resulting collaboration graph, the number of processed videos, the number of correctly and incorrectly detected collaborations, as well as the time needed for the calculations.
 
\subsection{Evaluation on the VoxCeleb dataset}
This part of the evaluation is based on a subset of the previously described VoxCeleb dataset. Per definition by VoxCeleb~\cite{Nagrani2017}, the videos should always contain the labeled person with a visible face and actively speaking. The videos thereby consist mostly of interviews and therefore could contain other people as well.\\
The results on this test set for CATANA are $86$\% correct assigned videos and $14$\% of incorrectly assignments. Results have a homogeneity of $0.87$, a completeness of $0.89$ and a v-measure $0.88$. The results are as expected, as in this dataset every speaker is visible in the video. Discrepancies can result through other individuals appearing in the videos, which are not labeled in the dataset, for example, the interviewer or reporter. Homogeneity and completeness score are both positive and show that the found clusters consist mostly of data from a single class and further that the classes are not scattered over multiple clusters but mainly contained in a single cluster each.\\
For the evaluation using our proposed extension, a pruning of the VoxCeleb test set had to be done due to the time consumption of processing all $700$ videos. The test set was thereby shortened to $8$ speaker and $40$ videos. Regarding evaluation comparison this is not ideal, but otherwise no result could have been obtained in time.
Results for our proposed approach came out worse than anticipated and actually performed poorer on this test set than CATANA.
We can report, that $60$\% have been assigned correctly and 40\% incorrectly. The homogeneity has a value of $0.58$, the completeness is $0.55$, and the v-measure is $0.55$. Homogeneity and completeness signify both that the data was scattered into multiple clusters showing that the contained classes could not be identified and separated correctly.
Possible reasons for these results could be the varying video and audio quality of the set. After examining parts of the videos, we found strong difference in quality for videos of the same individual, which could lead to no, or incorrect recognition especially for speakers.\\
The results are shown in Table~\ref{table:voxCelebComparison} for an easy comparison of the evaluated approaches on the VoxCeleb dataset.

\begin{figure}[b]
\includegraphics[width=0.95\columnwidth]{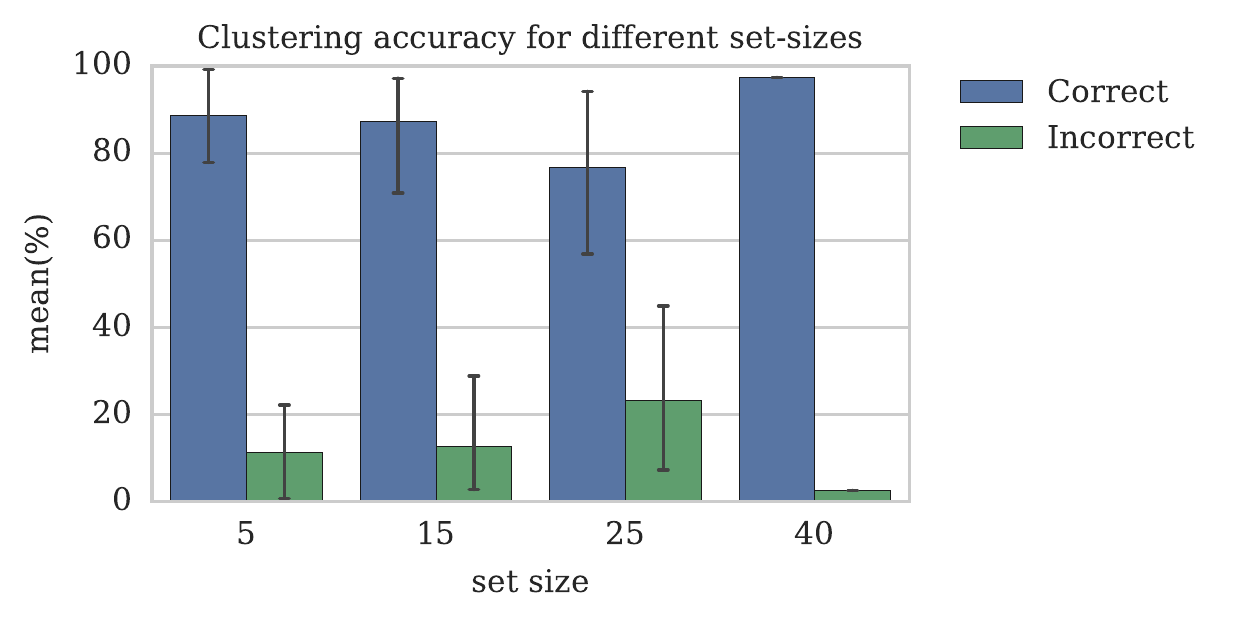}
\caption{Speaker clustering results for different set sizes from the VoxCeleb dataset.}
\label{fig:speaker_set_sizes_results}
\end{figure}

\begin{table*}[!t]
  \centering
  \begin{tabular}{ l | lllll}
    \toprule
     YouTube Gaming	   
     Collection &	Videos detected & Videos not detected & Collaborations found & Correct collaborations &  Time (h)\\ 
     \midrule
     Desired Values	& 	72		&   0  	& 	 34		&	34 & - \\
     
    CATANA 	& 	 31		&   41  	& 	 13		&	0	 &  6\\ 
    
    Proposed method &  63		& 	9	& 	29	& 8  & 55 \\
    \bottomrule
  \end{tabular}
	\caption{The results on the YouTube gaming video set for CATANA and our approach.} 
    \label{table:youTubeComparison}
\end{table*}

\subsection{Evaluation of speaker recognition only}
To further directly compare face and speaker recognition, we evaluated our proposed speaker recognition and diarization pipeline on the same VoxCeleb subset as previously while only utilizing the audio content.
The found results are $37\%$ correct assigned videos and  $63\%$ incorrect assignments. The results have a homogeneity of $0.43$, a completeness of $0.65$ and a v-measure of $0.51$. We can see that the speaker recognition alone performs significantly worse than face recognition and conclude that speaker recognition alone could not be a replacement for face recognition in this task. A reason for this result could also be the described difference in video and audio quality. A direct comparison can be found in Table~\ref{table:voxCelebComparison}.\\
To further evaluate the explanation for the worse performance of the speaker recognition, especially compared to their identification accuracy reported with $> 90\%$~\cite{Nagrani2017}, we conducted additionally tests on parts of the VoxCeleb dataset using their provided segment boundaries. This is done to ensure that only the labeled individual of the video is speaking in the audio. We therefore skipped speaker diarization which produced these segments instead in the first evaluation.
Tested using segments for $5$, $15$, $25$, and $40$ speakers we found that the accuracy significantly increased, nearing the reported accuracy in~\cite{Nagrani2017} with $> 90\%$. This indicates that the speaker diarization seems to have a big stake in the accuracy decrease of the first evaluation. Leading to the conclusion that the used speaker diarization system could not provide as good segmentations as the labeled data of the VoxCeleb dataset and seems to be a weak point in the pipeline.
Figure~\ref{fig:speaker_set_sizes_results} shows the mean results for the different set sizes using a confidence interval of $95$\%. Evaluation was thereby conducted in a $10$-fold fashion, selecting random individuals from a set of $40$ speakers for ten iterations. The narrow confidence interval for the set of $40$ speaker is due to that in every iteration all $40$ speakers of the test set were selected.

\subsection{Evaluation on gaming videos}
\begin{figure*}
\centering
\includegraphics[width=0.5\textwidth]{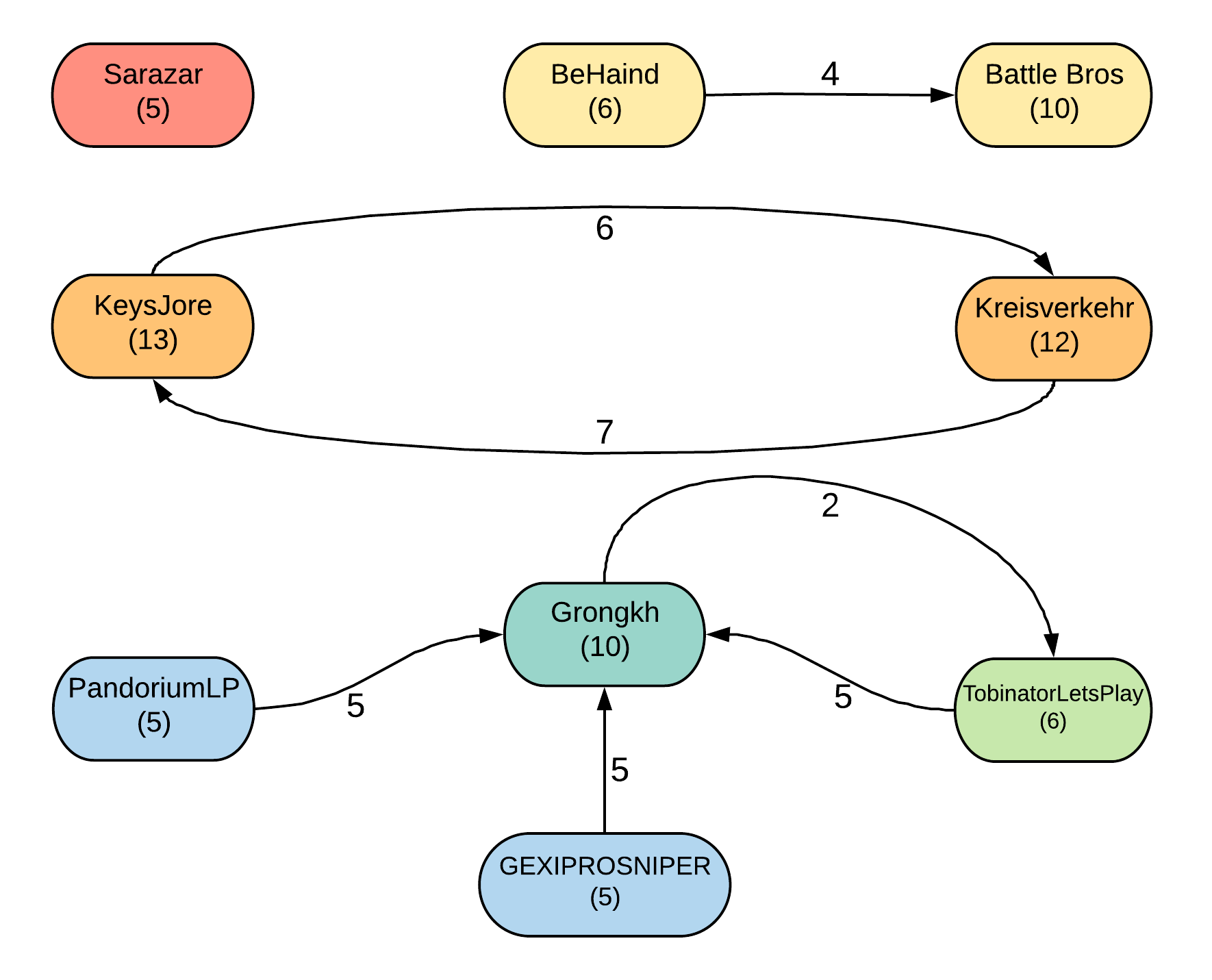}
\caption{Collaboration graph of the YouTube video test set. Showing channels as nodes, including channel name and number of videos in the set. The numbers on the edges represent the number of videos and collaborations between the channels.}
\label{fig:analyseSollGaming}
\end{figure*}

\begin{figure*}[!t]
\centering
\subfloat[CATANA detected collaborations.]{\includegraphics[width=0.95\columnwidth]{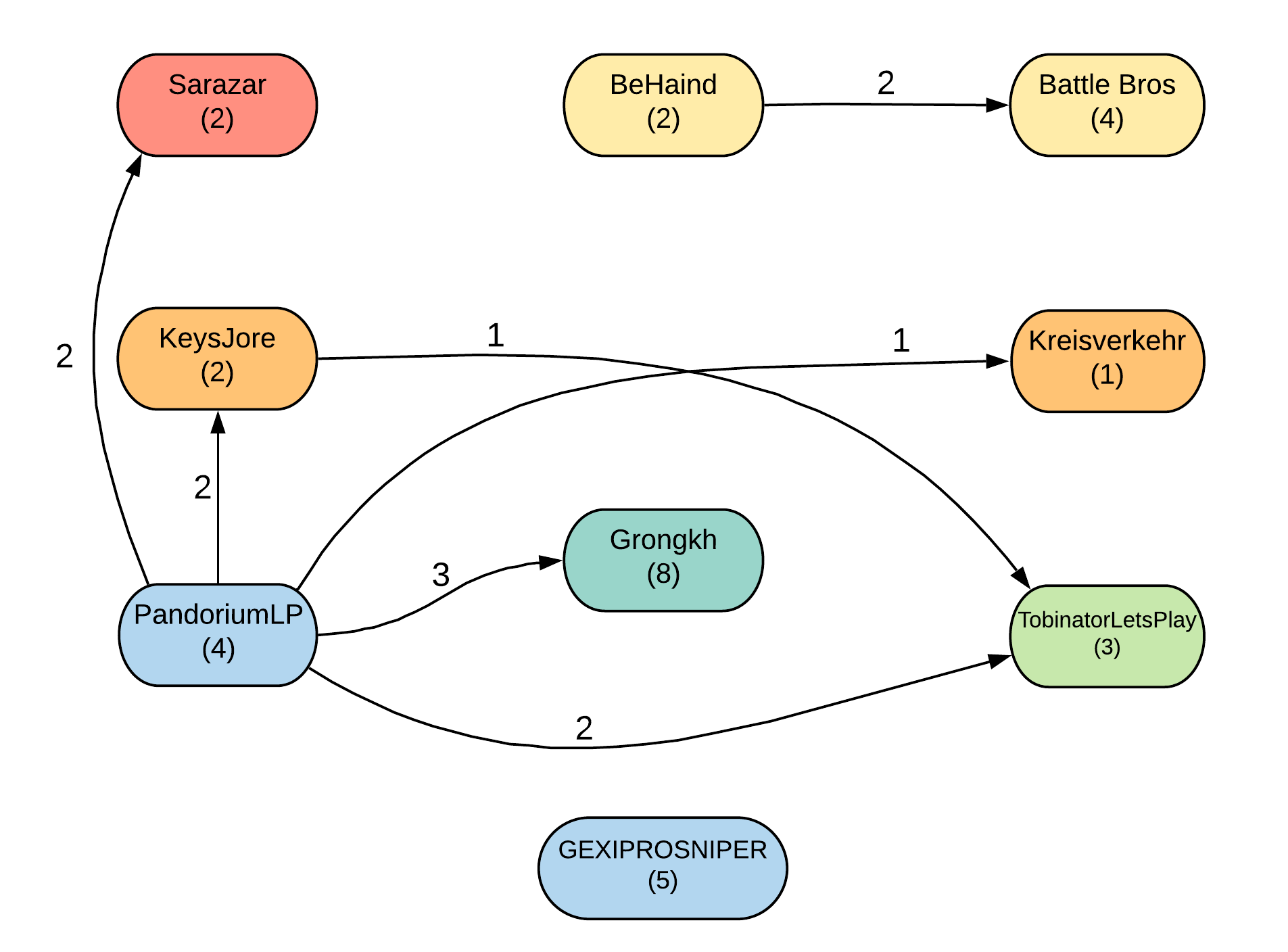}\label{fig:youtube_test_catana}}
\hfil
\subfloat[Proposed system detected collaborations.]{\includegraphics[width=0.95\columnwidth]{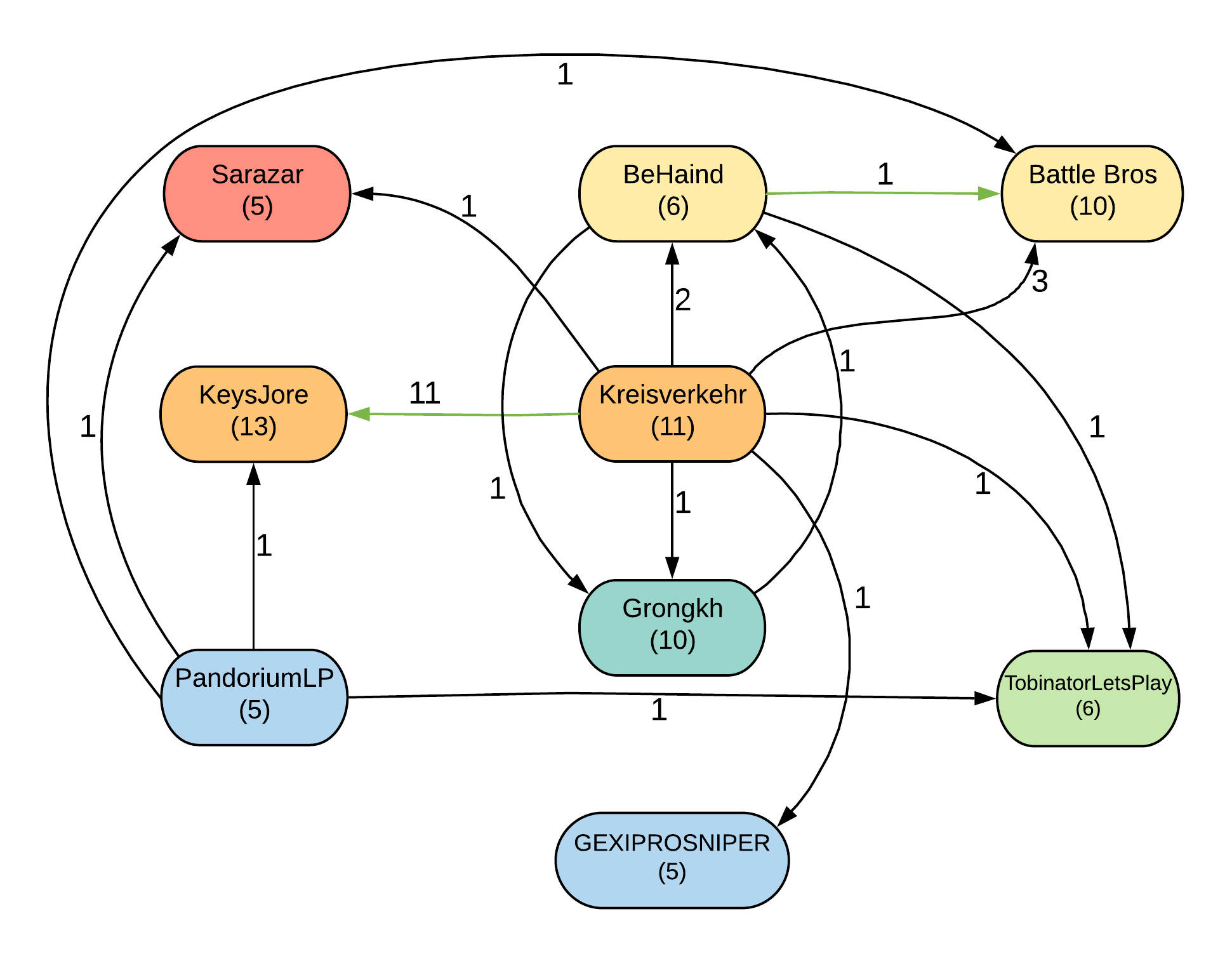}\label{fig:youtube_test_own}}
\caption{Collaboration results for CATANA and our new proposed approach. Correct collaboration edges marked green.}
\end{figure*}

In order to evaluate our proposed extension concerning the cases which CATANA failed to detect, we conducted an evaluation on a dataset consisting mainly of videos with no visible faces. The dataset contains $72$ videos from nine channels, of which $47$ videos contain only speech without appearing faces. A total of $34$ collaborations occur in the dataset, modeling a collaboration graph with $9$ nodes and $7$ edges displayed in Figure~\ref{fig:analyseSollGaming}.
The results of CATANA are, not surprisingly, very poor. CATANA was only able to extract and cluster embeddings from $31$ of the $72$ videos. In total $13$ collaborations were detected, resulting in a collaboration graph with $9$ nodes and $7$ edges.
For parts of the found collaboration-edges between the channels \emph{BeHaind} and \emph{Battle Bros} the detections were correct but falsely assigned the face of \emph{BeHaind} as the content creator of \emph{Battle Bros}. This lead to no edge in the collaborations graph from \emph{BeHaind} to \emph{Battle Bros}, but instead an edge from Battle Bros to \emph{BeHaind}. The Reason for this false assignment was the number of videos where \emph{BeHaind} appeared, which were higher in \emph{Battle Bros} than on his own channel. We suppose a larger dataset could prevent this error.
Concerning appearing faces in the dataset, in $82$\% of the videos detected through CATANA, faces actual appeared. In the remaining $17$\% of the videos, CATANA detected faces where no faces had appeared. Possible false detections could potentially be caused through appearing game characters.\\
The evaluation of our proposed extension results has improved the results but is not satisfying regarding accuracy and quantity.
Table~\ref{table:youTubeComparison} shows the results of the evaluation on the YouTube dataset. In comparison to CATANA's results, more videos could be detected and embeddings extracted. In total, we found $896$ embeddings which were clustered to $56$ clusters (not including noise). 
$41$ of these clusters were speaker clusters and $15$ were face clusters. The resulting collaboration graph contains $9$ nodes, $16$ edges and a total of $29$ detected collaborations. The collaboration graph is shown in Figure~\ref{fig:youtube_test_own}. In comparisons to CATANA, more collaborations were detected including an improvement in the number of correct detected collaborations.
We could notice that most of the collaborations including only two content creators were detected correctly. Other collaborations with up to $4$ participating content creators were not correctly detected. This could potentially be attributed to the previous results on speaker diarization concerning discussions with multiple people. Another point is an implementation detail concerning the execution of speaker diarization.
The initial design decision was to execute speaker diarization only after face tracking and active speaker detection, and if no active speaker was detected. Due to emerging cases with only a single active speaker combined with multiple non-visible speakers, thus decision reveals a flaw and could affect accuracy further. Adapting our design to this case is an easy task, however re-evaluating the YouTube dataset is not possible due to the time constraints of this work.
Another considerable factor is the time. While CATANA needed five hours to calculate the results, our proposed algorithm took $55$ hours, mainly attributable to the face tracking. The results are slightly better, for almost ten times the time CATANA needs.\\






\section{Discussion}
During the evaluation, we could determine that the implemented pipeline still has vast room for improvements.
Due to significant problems in development and acquiring satisfying results of individual pipeline parts,
we were not able to invest a lot of time to improve the accuracy and quality of the overall results of our approach.
Especially due to the time-consuming evaluation, as parts like face tracking are very time expensive. We suggest that CATANA is about ten times faster in processing videos than our approach. For this reason, the approach may be more suited to investigate small sets of videos rather than big sets, at least at this stage of development.

Video quality is also a major criterion for the output. In many YouTube formats especially Gaming, the content creator is visible on a so-called face cam, a small video stream of the face which is placed in one of the corners in the video. We determined, that only good results could be achieved for face cams and other small faces in the video if the video quality is at least 720p ($1280$x$720$ pixel). This ensures, that the cropped face has still a resolution of $160$x$160$. If the cropped face should be smaller, the image will be scaled up to the size of $160$x$160$. This results in a blurry face image leading to bad results in active speaker and face recognition.
Concerning our YouTube test set, videos showing game content had surprisingly no noticeable impact on the results. As it was initially thought that especially game characters could trigger speaker and face recognition, which however was not the case in our results.

\section{Conclusion}
The goal of this work was to design and implement an approach for the detection of content creator collaborations in YouTube videos. We based our approach on the existing work of CATANA~\cite{CATANA} and proposed an extension, which makes use of face tracking, active speaker detection and speaker recognition. This extension was designed to address open issues with CATANA's face recognition approach concerning videos without appearing faces. 
We implemented a working pipeline which extracts face tracks from videos, decides whether an individual is currently speaking and labels active speakers. Speech is extracted and an embedding is calculated for clustering occurring speakers. 
Additionally, the existing face recognition method is applied as well. In the best case, face and speaker embeddings describing the same individual are extracted and associated, allowing us to recognize this individual separate through either speech or face features.
Evaluation of the implemented system was conducted on two datasets and compared to results of CATANA.
Results for the collaboration detection in gaming videos with mainly no visible faces showed an improvement in accuracy and quantity but also flaws in the current pipeline. One weak point of the pipeline being the speaker diarization.\\
Therefore, we conclude that the proposed approach has potential in the application of collaboration detection but needs further refinement to make it a viable replacement for CATANA and beyond.


%





\ifCLASSOPTIONcaptionsoff
  \newpage
\fi

\end{document}